\documentclass[letterpaper, 10 pt, conference]{ieeeconf}

\usepackage{gensymb}
\usepackage{hyperref}
\usepackage{tabularx}
\usepackage{graphicx}
\usepackage{caption}
\usepackage{subfigure}
\usepackage{multirow}
\usepackage{amsmath}
\usepackage{array}
\usepackage{pifont}
\usepackage[usenames, dvipsnames]{color}
\usepackage{cite}
\usepackage{subfigure}
\usepackage{graphicx}
\usepackage{xcolor}
\usepackage{array,booktabs}
\usepackage{tabularx}
\IEEEoverridecommandlockouts

\title{\LARGE \bf
Dynamic Traffic Scene Classification with Space-Time Coherence}
\author{Athma Narayanan$^{1*}$, Isht Dwivedi$^{2*\dagger}$, Behzad Dariush$^{1}$% <-this % stops a space
\thanks{$^{1}$Honda Research Institute USA, Mountain View, CA 94043, USA
{\tt\small \{anarayanan, bdariush\}@honda-ri.com}}%
\thanks{$^{2}$Department of Computer Science, Columbia University, NYC, NY 10027  , USA. 
{\tt\small isht.dwivedi@columbia.edu}}%
\thanks{$^{*}$co-first authors who contributed equally to this work}%
\thanks{$^{\dagger}$This work was done while Isht Dwivedi was an intern at Honda Research Institute USA}%
}

\begin{document}

\setlength{\skip\footins}{0.2cm}
\maketitle
\thispagestyle{empty}
\pagestyle{empty}

%%%%%%%%%%%%%%%%%%%%%%%%%%%%%%%%%%%%%%%%%%%%%%%%%%%%%%%%%%%%%%%%%%%%%%%%%%%%%%%%
\begin{abstract}
This paper examines the problem of dynamic traffic scene classification under space-time variations in viewpoint that arise from video captured on-board a moving vehicle.  Solutions to this problem are important for realization of effective driving assistance technologies required to interpret or predict road user behavior.  Currently, dynamic traffic scene classification has not been adequately addressed due to a lack of benchmark datasets that consider spatiotemporal evolution of traffic scenes resulting from a vehicle's ego-motion.    This paper has three main contributions. First, an annotated dataset is released to enable dynamic scene classification that includes 80 hours of diverse high quality driving video data clips collected in the San Francisco Bay area.  The dataset includes temporal annotations for road places, road types, weather, and road surface conditions.   Second, we introduce novel and baseline algorithms that utilize semantic context and temporal nature of the dataset for dynamic classification of road scenes.   Finally, we showcase algorithms and experimental results that highlight how extracted features from scene classification serve as strong priors and help with tactical driver behavior understanding. The results show significant improvement from previously reported driving behavior detection baselines in the literature.
\end{abstract}

%%%%%%%%%%%%%%%%%%%%%%%%%%%%%%%%%%%%%%%%%%%%%%%%%%%%%%%%%%%%%%%%%%%%%%%%%%%%%%%%
\section{INTRODUCTION}

Semantic description and understanding of dynamic road scenes from an egocentric video is a central problem in realization of effective driving assistance technologies required to interpret and predict road user behavior.   In the driving context, scene refers to the place where such behaviors occur, and includes attributes such as environment (road types), weather, road-surface, traffic, lighting, etc.   Importantly, scene context features serve as important priors for other downstream tasks such as recognition of objects, behavior, action, intention, as well as robust navigation, and localization.  For example, cross-walks at intersections are likely places to find pedestrians crossing or waiting to cross.   Likewise, knowing that an ego-vehicle is approaching an intersection helps auxiliary modules to look for traffic lights to slow down.  Needless to say, effective solutions to the traffic scene classification problem provide contextual cues that promise to help driving assist technologies to reach human level visual understanding and reasoning.

\begin{figure}
\includegraphics[height=6cm,width=\linewidth]{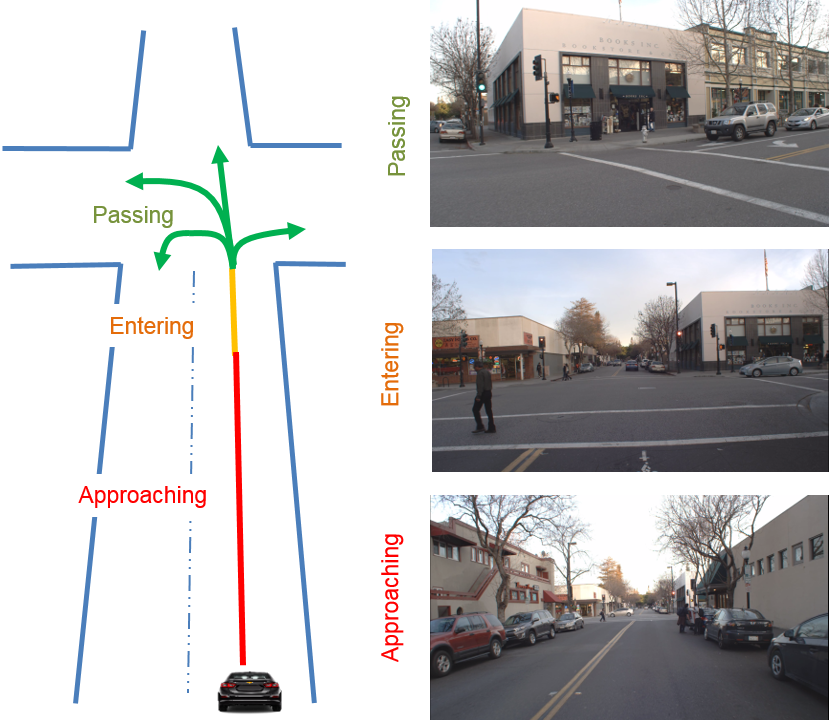}
\centering
\caption{Viewpoint variations when approaching, entering, and passing through an intersection.}
\label{fig:Intro_AEP}
\end{figure}

\begin{figure*}
\centering 
\includegraphics[scale=0.5,width=\textwidth]{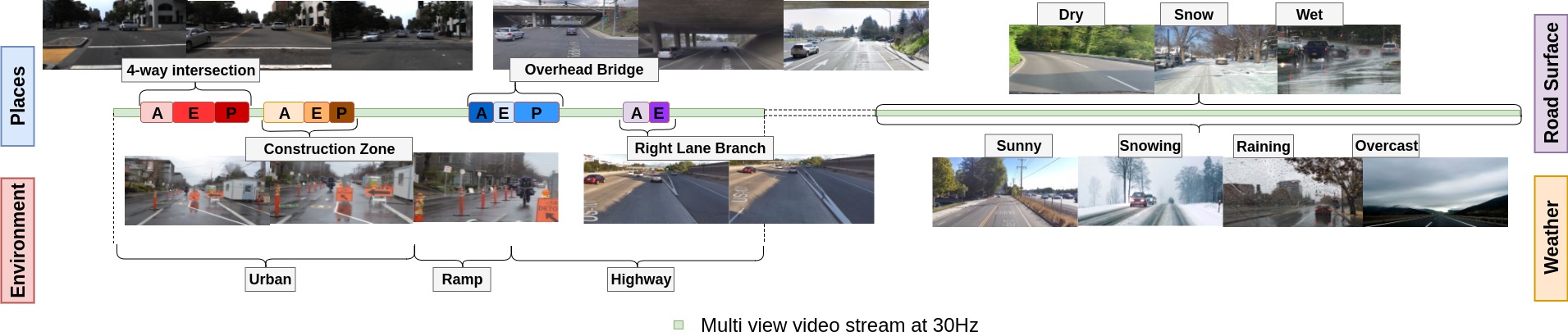}
\caption{Temporal video annotations at multiple levels, including Places, Environments, Weather, and Road Surface. 
}
\centering
\label{fig:intro}
\end{figure*}

The vast majority of research in scene classification has been conducted to address the problem of single image classification~\cite{zhou2017places}~\cite{yu2015lsun}.   Recently, dynamic scene classification datasets~\cite{shroff2010moving} and associated algorithms~\cite{feichtenhofer2016dynamic} have emerged that exploit spatiotemporal features.   
However, majority of previous work consider a stationary camera and study image motion (i.e. spatial displacement of image features) that is induced by projected movement of scene elements over time.   Typical examples include a rushing river, waterfall, or motion of cars on the highway from a surveillance camera.    For driving tasks, scene understanding requires a dynamic representation characterized by displacement of image motion attained from moving traffic participants as well as variations in image formation that emerge from the vehicle’s ego-motion.    The latter is an important and challenging problem that has not been addressed, primarily due to a lack of related datasets for driving scenes.

To address this solution gap, this paper critically examines the dynamic traffic scene classification problem under space-time variations in viewpoint (and therefore scene appearance) that arise from the egocentric formation of images collected from a moving vehicle.    In particular, a novel driving scene video dataset is introduced to enable dynamic traffic scene classification.    The dataset includes temporal annotations on place, environment (road-type), and weather/surface conditions and explicitly labels the viewpoint variations using multiple levels.    Specifically, the place categories are annotated temporally with fine grained labels such as Approaching (A), Entering (E), and Passing (P), depending on the ego-car’s relative position to the place of interest.    An example of this multi-level temporal annotation is depicted in Figure~\ref{fig:Intro_AEP} and ~\ref{fig:intro}.   This example illustrates the result of view variations (caused by changing distance to the intersection) as a vehicle approaches the scene of interest (i.e. intersection).   The video clip is labeled using the three layers (A,E,P) to highlight the distinct appearance changes and showcases the proposed fine grained annotation strategy that is important for vehicle navigation and localization.

The main contributions of this work are as follows.   First, a dataset is released that includes 80 hours of diverse high quality driving video data clips collected in San Francisco Bay area
\footnote{\url{https://usa.honda-ri.com/hsd}}. The dataset includes temporal annotations for road places, road environment, weather, and road surface conditions. This dataset is intended to promote research in fine-grained dynamic scene classification for driving scenes.   The second contribution includes development of machine learning algorithms that utilize the semantic context and temporal nature of the dataset to improve classification results.    Finally, we present algorithms and experimental results that showcase how extracted features can serve as strong priors and help with tactical driver behavior understanding.

\begin{figure*}
\includegraphics[height=6cm,width=\linewidth]{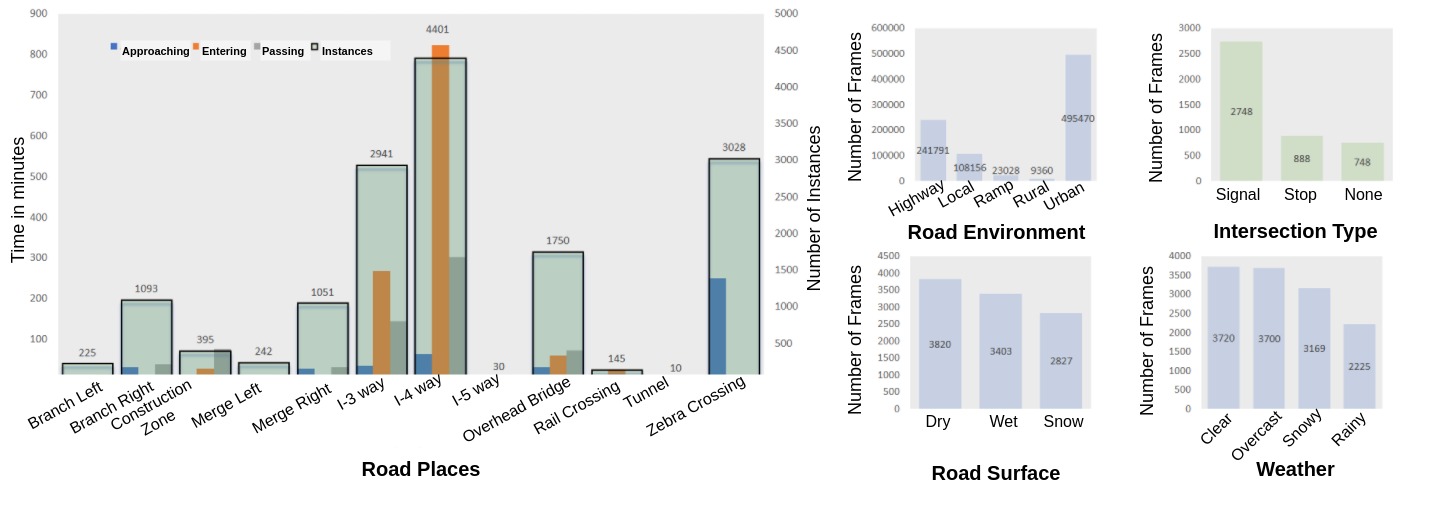}
\centering
\caption{Left bar plot shows the fine grained distribution of Road Places, where ``I'' denotes intersection.  Right bar plots depict other classes supported by our dataset. Top left and bottom row shows the number of frames used to benchmark classification algorithms on Road Environment, Road Surface and Weather. The top right plot shows each 3-way, 4-way or 5-way intersection  labelled as an intersection with traffic signals, stop signs, or with none of the above.}
\label{placechart}
\end{figure*}

\section{RELATED WORK}
\subsection{Driving Data sets}
Large scale public datasets geared towards automated or advanced driver assist systems ~\cite{ geiger2012we, jain2015car, chen2018lidar, xu2017end, ramanishka2018toward, yu2018bdd100k, maddern20171}, and scene understanding ~\cite{huang2018apolloscape, cordts2016cityscapes, madhavan2017bdd,neuhold2017mapillary}, have helped the development of algorithms to better understand the scene layout and behavior of traffic participants.  These datasets have limitations since they either do not adequately support dynamic scene classification or provide only a non exhaustive list of  driving scene classes.
Several papers support pixel wise annotations for semantic segmentation~\cite{madhavan2017bdd,huang2018apolloscape,neuhold2017mapillary,cordts2016cityscapes}. While it may be useful to learn semantic segmentation models to parse the scene, we cannot infer the type of scene reliably. This would mean that separate models need to be developed to aggregate the semantic segmentation outputs and infer the type of scene.   Other datasets provide models for understanding ego and participant behaviors in driving scenes~\cite{jain2015car,ramanishka2018toward}, but they do not have an exhaustive list for scene classes.   With respect to datasets, most similar to our work is described in~\cite{maddern20171,garg2018don't}.  While they provide labels for scene and weather classification, the dataset is more focused toward image retrieval and localization problems. 
%We would like to build upon this work and provide a more exhaustive set of driving %scene labels that will serve as a scene context prior for other tasks.

\subsection{Scene and Weather classification}
MIT Places dataset~\cite{zhou2017places} and Large Scene understanding dataset (LSUN) ~\cite{yu2015lsun} were introduced to benchmark several deep learning based classification algorithms.  While our data set serves a similar purpose  , but for traffic scenes, we also support temporal classification to benchmark algorithms that are robust to spatio-temporal variations of the same scene.   Moreover driving scenes have an unbalanced class distribution along with less inter class variation, making classification much more challenging. For example, fine grained classification of 3-way and 4-way intersection from a single frontal camera view is very challenging due to small scene variations between these two classes.  

Existing frame based scene and weather classification can be grouped into the following methods: adding semantic segmentation and contextual information \cite{yao2012describing,li2017multi,lin2017rscm}, using  hand crafted features \cite{bolovinou2013dynamic,feichtenhofer2016dynamic,sikiric2014image}, multi resolution features \cite{wang2017knowledge,wu2017traffic}, or use multiple sensor fusion \cite{jonsson2011road,jonsson2015road}.  Given the success and superior deep learning classification methods, we elected to use a learning based approach along with experimentation on how to add semantic segmentation and temporal feature aggregation to improve the results.

\begin{table}[]
\centering
\setlength{\tabcolsep}{0.1pt}
\begin{tabular}{|c|c|c|c|c|}
\hline
\textbf{Datasets} & \textbf{Temporal} & \textbf{Purpose} & \textbf{Areas} & \textbf{Road} \\ \hline
 Cityscapes\cite{cordts2016cityscapes} & N & Sem. Segmentation & U & Y \\ \hline
 Appolscape\cite{huang2018apolloscape} & Y & Sem. Segmentation & U & Y \\ \hline
 BDD-Nexar \cite{madhavan2017bdd} & N & Sem. Segmentation & U, H & Y \\ \hline
 Kitti\cite{geiger2012we} & Y & Detection & U & Y \\ \hline
 DR(eye)VE \cite{dreyeve2018} & Y & Driver Behavior & U & Y \\ \hline
 HDD\cite{ramanishka2018toward} & Y & Driver Behavior & U,H,L & Y \\ \hline
 LSUN\cite{yu2015lsun} & N & Scene Understanding & - & N \\ \hline
 Places\cite{zhou2017places} & N & Scene Understanding & - & N \\ \hline
 Ours & Y & Scene Understanding & U,H,L,R & Y \\ \hline
\end{tabular}
\caption{Comparison of datasets. U-\textit{Urban}, H-\textit{Highway}, L-\textit{Local}, R-\textit{Rural}, Temporal - \textit{Temporal annotations}, Road- \textit{Traffic scenes}} 
\label{tablecomp}
\end{table}

% \begin{table}[]
% \setlength{\tabcolsep}{1pt}
% \renewcommand{\arraystretch}{1.5}
% \centering
% %\resizebox{\linewidth}{!}{
% \begin{tabularx}{|c|c|c|c|c|}
% \hline
%  \textbf{Datasets} & \textbf{Temporal} & \textbf{Purpose} & \textbf{Areas} & \textbf{Road scene} \\ \hline
%  Cityscapes\cite{cordts2016cityscapes} & No & Semantic segmentation & Urban & Yes \\ \hline
%  Appolscape\cite{huang2018apolloscape} & Yes & Semantic segmentation & Urban & Yes \\ \hline
%  BDD-Nexar\cite{madhavan2017bdd} & Yes & Semantic Segmentation & Urban, Highway, Residential & Yes \\ \hline
%  Kitti\cite{geiger2012we} & Yes & Traffic Participant behavior & Urban & Yes \\ \hline
%  DR(eye)VE & Yes & Driver behavior & Urban & Yes \\ \hline
%  HDD\cite{ramanishka2018toward} & Yes & Driver behavior & Urban, Highway, Residential & Yes \\ \hline
%  LSUN\cite{yu2015lsun} & No & Scene Understanding & - & No \\ \hline
%  Places\cite{zhou2017places} & No & Scene Understanding & - & No \\ \hline
%  Ours & Yes & Scene Understanding & Urban, Highway, Residential,Rural & Yes \\ \hline
% \end{tabularx}%
% %}
% \caption{Comparison of datasets}
% \label{tablecomp}
% \end{table}

\subsection{Temporal Classification}
Video classification and human activity recognition tasks ~\cite{karpathy2014large,tran2015learning,feichtenhofer2016convolutional}  have helped develop various state of the art deep learning methods for temporal aggregation. These methods aggregate spatio-temporal features through Long short-term memory modules (LSTM)~\cite{yue2015beyond} or Temporal Convolution Networks (TCN) ~\cite{r2plus1d_cvpr18}. While such methods help activity recognition tasks by understanding object level motion primitives, they do not translate directly for temporal scene classification. In fact, frame based result averaging might be more suitable for our problem.  Moreover, the entire scene is the focus of our task, not just the human actor.

Recently, work has been done for region proposal generation~\cite{chao2018rethinking,xu2017r}, where two stream architectures are used to generate the start and end time of the event as well as the class activity. Our work is inspired by these methods. Our best model is a two stream architecture that decouples the region proposal and classification tasks.  Specifically, we use the proposal generator to trim the untrimmed video and aggregate the features to classify the entire trimmed segment. This method outperforms simple frame based averaging techniques. For example, it is better to come to conclusion if the class is a 4-way intersection by looking at the segment (approaching, entering and passing) in its entirety rather than on a per frame basis. This helps the model parse the same intersection from various viewpoints.  Details of the proposed method is provided in Section IV.

% \begin{figure}
% \centering
% \subfigure[]
% {\includegraphics[width=5cm]{Images/carsetup.png}}
% \subfigure[]
% {\includegraphics[width=3cm,height=2.3cm]{Images/gmaps.jpg}}
% \qquad
% \caption
% {(a) The data collection platforms.  The first vehicle contains 3 NVIDIA RCCB(60FOV) and 1 NVIDIA RCCB(100FOV). The second vehicle contains 2 Pointgrey Grasshopper (80FOV) and 1 Grasshopper (100FOV). (b) Recording paths around the San Francisco Bay Area.}
% \label{carsetup}
% \end{figure}

\section{Overview Of the Honda Scenes Dataset}
\label{dataset}
\subsection{Data collection platform}
Our collection platform involves two instrumented vehicles, each with different camera hardware and setup. The first vehicle contains \textit{three} NVIDIA RCCB(60FOV) and \textit{one} NVIDIA RCCB(100FOV). The second vehicle contains \textit{two} Pointgrey Grasshopper (80FOV) and \textit{one} Grasshopper (100FOV). This varied setup enables development of algorithms that support better generalization to camera hardware and positioning. The cameras cover approximately 160 degree frontal view. 
%The first vehicle is a data collection vehicle with cameras mounted inside the windshield.  The second vehicle has outdoor cameras mounted on the roof of the vehicle.  The cameras cover approximately 160 degree frontal view. 

Data was collected around the San Francisco Bay Area region over a course of six months under different weather and lighting conditions. Urban, Residential(Local), Highway, Ramp, and Rural areas are covered in this dataset.  Different routes are taken for each recording session to avoid overlap in the scenes. Moreover, targeted data collection is done to reduce the impact of unbalanced class distribution for rare classes such as railway crossing or rural scenes. The total size of the post-processed dataset is approximately 60 GB and 80 video hours. The videos are converted to a resolution of $1280 \times 720$ at $30$ fps.

\subsection{Data set statistics and comparison}
Table~\ref{tablecomp} shows the overall comparison with other data-sets.  Our dataset is the only large scale driving dataset for the purpose of driving scene understanding. The datasets were annotated with exhaustive list of classes typical to driving scenarios. Three persons annotate each task and cross-check results to ensure quality. Intermediate models are trained to check the annotations and to scale the dataset with human in loop.    The ELAN \cite{brugman2004annotating} annotation tool is used to annotate videos at multiple levels.   The levels include Road Places, Road Environment (Road Types), Weather, and Road Surface condition.  The data is split into training and validation in such a way so as to avoid any geographical overlap. This enforces generalization of models to new unseen areas, changes in lighting condition, and changes in viewpoint orientation. Further details about the class distribution are described below.

\textbf{Road Places and Environment} The 80 hour video clips are annotated with Road Place and Road Environment labels in a hierarchical and in a causal manner. There are three levels in the hierarchy. At the top level, Road Environment is annotated, followed by the Road Place classes at the mid level, and the fine grained annotations such as \textit{approaching}, \textit{entering}, \textit{passing} at bottom level. This forms a descriptive dataset that allows our algorithms to learn the inter dependencies between the levels.
The Road Environment labels include  \textit{urban}, \textit{local}, \textit{highway} and \textit{ramps}.  The \textit{local} label includes residential scenes which are typically less traffic prone and contain more driveways as opposed to  \textit{urban} scenes.  The \textit{Ramps} class generally appear at highway exits and are connectors between two highways or a highway and other road types. 

Each fine grained annotation is clearly defined based on the view from the ego-vehicle. The \textit{three-way}, \textit{four-way}, and \textit{five-way} intersections each have \textit{approaching}, \textit{entering}, and \textit{passing} labels based on the ego-vehicle's position from the stop-line, traffic signal and or stop sign. Similarly \textit{construction zones}, \textit{rail crossing}, \textit{overhead bridge} and \textit{tunnels} are labelled based on the ego-vehicle's position from the construction event, railway tracks, overhead bridge, and tunnels, respectively.  Since the notion of entering does not exist or is too abrupt for \textit{lane merge, lane branch or zebra crossing} classes, these categories are annotated with only \textit{approaching} and \textit{passing} fine grained labels. The overall class distribution in illustrated in Figure~\ref{placechart}.

\textbf{Weather and Road Surface condition} Due to the lack of snow weather conditions in the San Francisco Bay Area, a separate targeted data collection was performed in Japan specifically for snow weather and snow surface conditions. This also helps the weather and surface prediction models generalize well to different places and road types. 
% In addition to this, we also use some selected images from the BDD100k dataset \cite{yu2018bdd100k} for experiments on \textit{road surface condition} and \textit{road weather}. The BDD100k dataset images has%
Video sequences are semi-automatically labeled using weather data and GPS information before further processing and quality checking by human annotators.  

The temporal annotations for weather contain classes such as \textit{clear}, \textit{overcast}, \textit{snowing}, \textit{raining} and \textit{foggy} weather conditions. The Road Surface has \textit{dry}, \textit{wet}, \textit{snow} labels. Only frames with sufficient snow coverage on the road (more than 50\%) are labeled as snow surface condition.
This maintains the road surface and the weather condition labels to be mutually exclusive to each other.
%  There are no openly available data-sets for road-surface condition estimation with sufficient number of images to train a deep CNN. \cite{qian2016evaluating} present a dataset which has only $100$ images. 
% Therefore, we built our own larger road surface dataset manually using images from our data and BDD100k datasets. 
% Our surface condition dataset has the \textit{dry}, \textit{wet} and \textit{snowy} surface classes. 
While we do provide temporal annotations, only sampled frames are used for our experiments. When predicting the conditions on untrimmed test videos, the results are averaged over a temporal window as these conditions do not change drastically frame to frame.  Figure~\ref{placechart} shows the distribution of images over classes for weather and road surface conditions. 

\section{METHODOLOGY}
This section describes the proposed methods for dynamic road scene classification for holistic scene understanding with respect to an ego vehicle driving on a road. Our proposed methods are able to predict \textit{road weather}, \textit{road surface condition}, \textit{road environment} and \textit{road places}.

\subsection{Experiments}
All experiments are based on the \textit{resnet50} model. It should be noted that that the proposed methods can be applied using any base Convolutional Neural Network (CNN). Any performance improvement on the base CNN could directly transfer to performance improvement on our methods.  These models run on the NVIDIA P100 at 10 \textit{fps}.

\begin{figure}
\centering
\includegraphics[width=4cm,height=1.9cm]{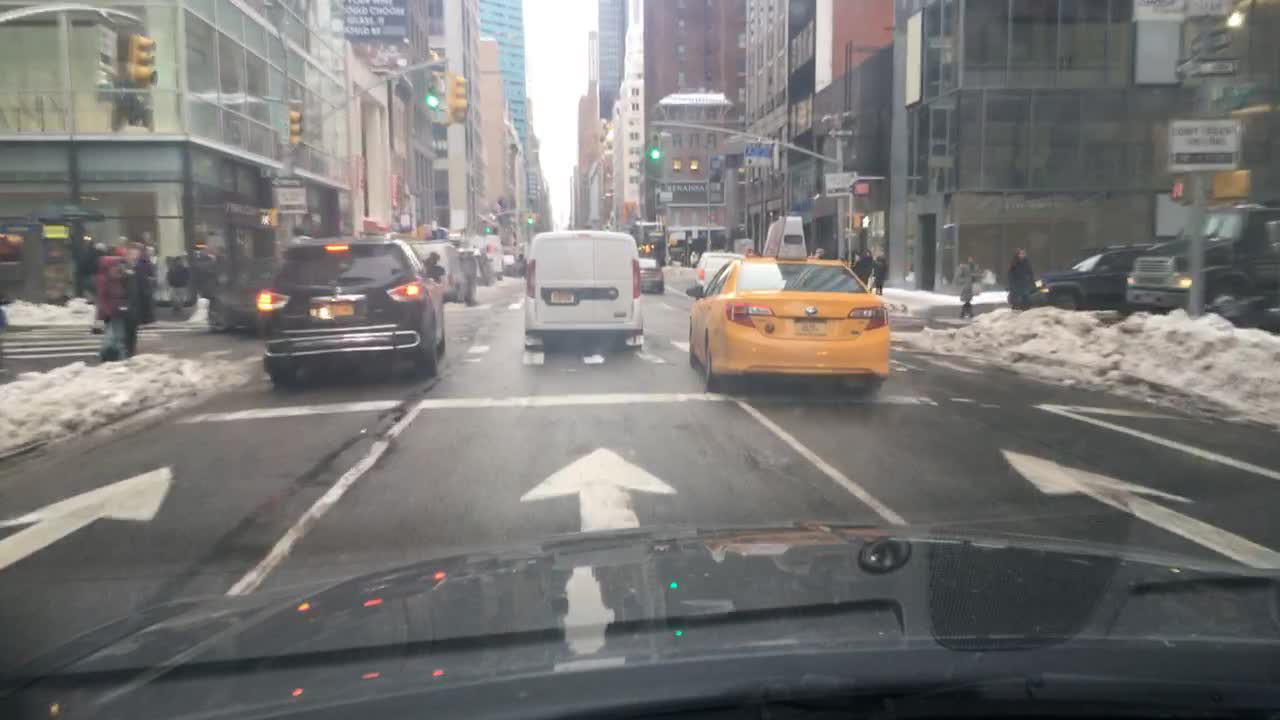}
\includegraphics[width=4cm,height=1.9cm]{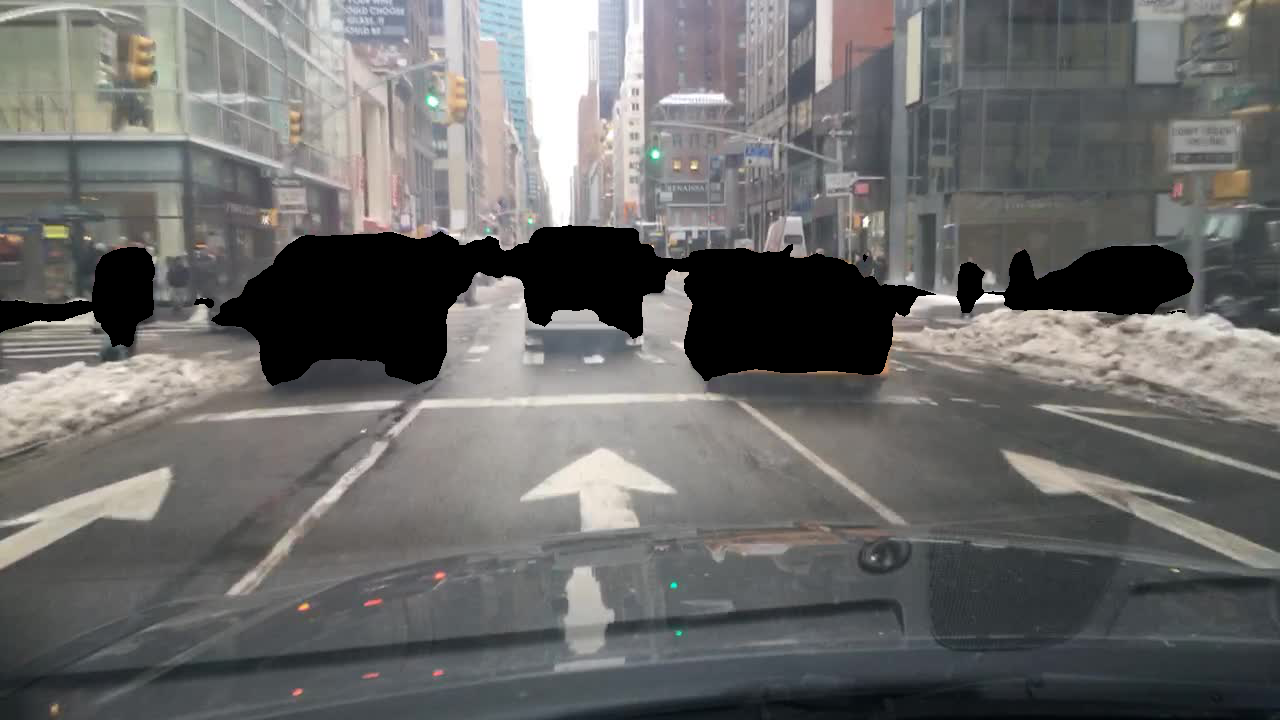}
\qquad
\caption
{A sample image from our training set and the corresponding RGB-masked image where the traffic participants are removed.}
\label{sampleimage}
\end{figure}

\textbf{Road Weather and Road Surface Condition: } To classify weather and road surface, we chose to train a frame based \textit{resnet50}~\cite{he2016deep} model. For weather and road surface, approximately \textit{$3000$} images of each class were used to fine-tune models pre-trained on the places365~\cite{zhou2017places} dataset.   Since \textit{foggy} weather is a rare class, it was not used in our current set of experiments to avoid an unbalanced class distribution. As a first experiment, we finetuned a \textit{resent50} pretrained on the places365 dataset. 
% The dataset we obtained for this task after pruning BDD100k has  and \textit{foggy} weather classes.
The weather category is independent of traffic participants in the scene.  Therefore, the base model \textit{resnet50} was fine-tuned on images where traffic participants were masked out as shown in Figure~\ref{sampleimage}. A semantic segmentation model based on Deeplab~\cite{chen2018deeplab} was used to segment and mask the traffic participants and allow the model to focus on the scene. 
% In another experiment, we trained \textit{resnet50} on a 4 channel image, where the $4^{th}$ channel is the semantic segmentation output map. Additionally for road surface we also 
The results are presented in Table~\ref{weathersurfacetable}, illustrating that semantic masking improves performance. 

\begin{table}[t]
\begin{center}
\resizebox{\linewidth}{!}{
\begin{tabular}{|*{10}{c|}}  % repeats {c|} 18 times
\hline
\multicolumn{1}{|c}{Input} &
\multicolumn{5}{|c}{Weather} & \multicolumn{4}{|c|}{Road Surface} \\ \hline
\hline
% \multicolumn{3}{|c}{50 clusters} & \multicolumn{3}{|c}{60 clusters} & \multicolumn{3}{|c}{70 clusters} & 
% \multicolumn{3}{|c}{50 clusters} & \multicolumn{3}{|c}{60 clusters} & \multicolumn{3}{|c|}{70 clusters} \\ \hline 
 -& clear & overcast & rain & snow & \textbf{mean} & dry & wet & snow & \textbf{mean} \\ \hline
 \hline
 
 RGB &  0.86 & 0.83  & 0.83  & 0.82  & 0.83 &  0.92 & 0.90  & 0.992  & 0.94 \\ \hline
 RGB\textit{(masked)} & \textbf{0.92}  & 0.83  & \textbf{0.96}  & \textbf{0.94}  &\textbf{0.91 } & \textbf{0.93}  &\textbf{0.92}   & \textbf{0.997}  &\textbf{0.95}  \\ \hline
 %RGBS &   &   &   &   &  &   &   &   &  \\ \hline

%  & & & & & & & & & & & & & & & & &  \\ \hline
\end{tabular}}
\caption{class-wise F-Scores for weather classification and road surface condition classification models}
\label{weathersurfacetable}
\end{center}
\end{table}

\begin{table}[t]
\begin{center}
%\tiny{
\resizebox{\linewidth}{!}{
\begin{tabular}{|*{6}{l|}}  % repeats {c|} 18 times

\hline
Input & Highway & Urban &  Local & Ramp & \textbf{mean}\\ \hline \hline 
 RGB & 0.86 & 0.81 & 0.33 & 0.07 & 0.52\\ \hline
 RGB\textit{ (masked)} & \textbf{0.91} & \textbf{0.83} & 0.33 & 0.20 & \textbf{0.56}\\\hline
 RGBS\textit{ (4 channel)}  & 0.89 & 0.81 & \textbf{0.34} & 0.13 & 0.54 \\\hline
 S\textit{ (1 channel)}& 0.90 & 0.81 & 0.24 & \textbf{0.25} & 0.55\\ \hline
 %RGBS &   &   &   &   &  &   &   &   &  \\ \hline

%  & & & & & & & & & & & & & & & & &  \\ \hline
\end{tabular}
%}
}
\caption{Class-wise F-scores for road environment }
\label{roadtypetables}
\end{center}
\end{table}

% \textbf{ Road Surface: } Our surface condition dataset has the \textit{dry}, \textit{wet} and \textit{snowy} surface classes with about $\sim3000$ images for each class. Similar to road weather classification, we fine-tuned resnet50 with RGB and RGBS inputs in 2 separate experiments. Some previous works \cite{amthor2015road} in road surface condition estimation use fixed rectangular patches of road area for road surface condition estimation. Therefore, in one experiment, we also used such rectangular patches as inputs to fine-tune resnet50. As an additional experiment, we used the semantic segmentation map to mask out non-road parts from the input image  and then only this masked image was used to fine-tune our model.

\textbf{ Road Environment: } For Road Environment, experiments were performed with \textit{resnet50}  pre-trained on places365 dataset. Similar to weather and road surface experiments the input to the model was progressively changed with no change to the training protocol. More specifically, experiments were conducted on the original input images (RGB), images concatenated with semantic segmentation (RGBS), images with traffic participants masked using semantic segmentation (RGB-masked), and finally only using a one channel semantic segmentation image (S). The class wise results are shown in Table \ref{roadtypetables}. 

Interestingly, while RGB-masked images show overall best performance, semantic segmentation alone outperforms just RGB images, especially for \textit{ramp} class. This might be due to the fact that scene structure is sufficient to understand the curved and the uncluttered nature of highway ramps. However, while decomposing the images to scene semantics allows the model to learn valuable structure information, it loses important texture cues about the type of buildings and driveways. Hence there is a lot of confusion between \textit{local} and \textit{urban} class resulting in lower \textit{local} performance. 

\begin{figure*}
\includegraphics[scale=0.21,trim={0.6cm 0 0.6cm 0.2cm},clip]{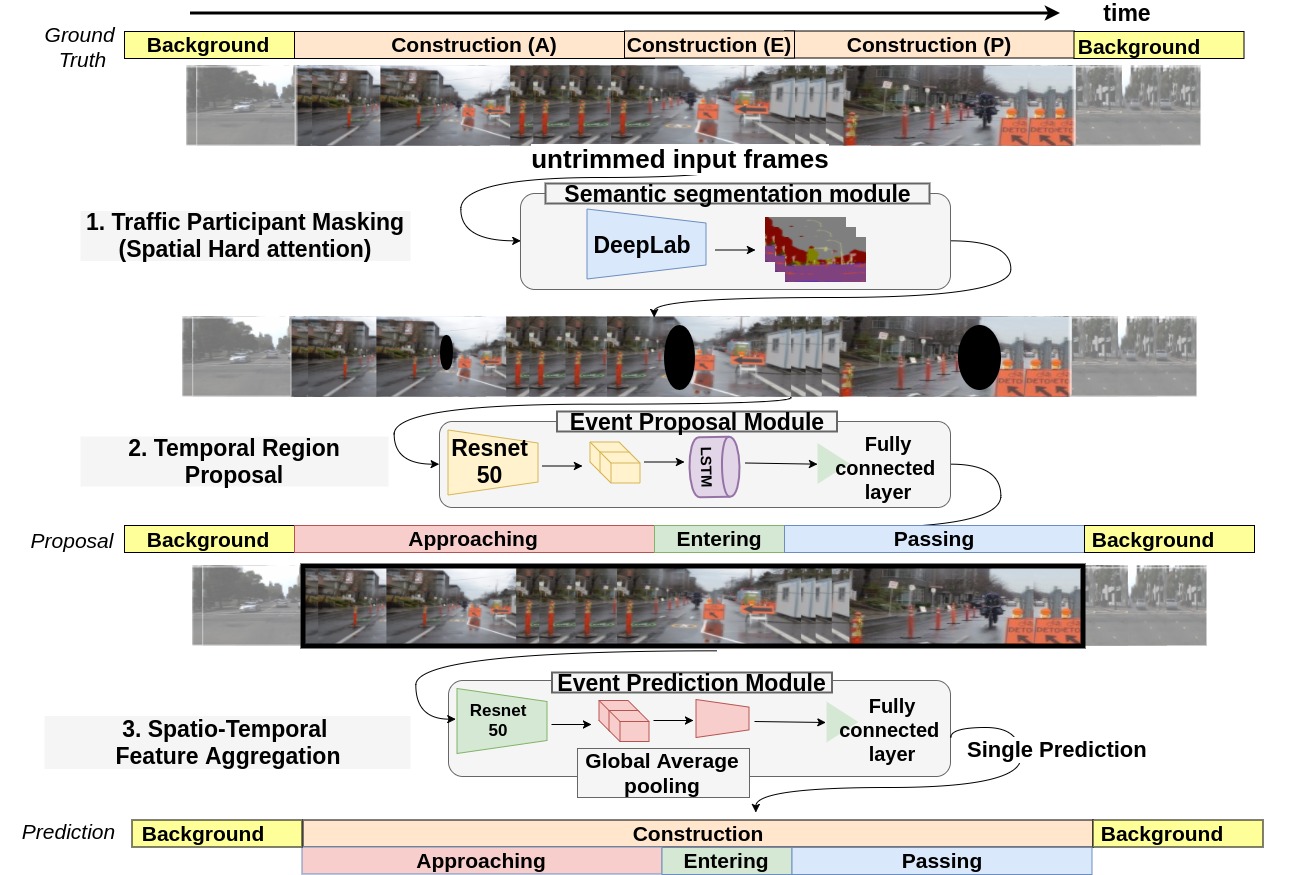}
\centering
\caption{Event Proposal Outline}
 \label{places_block}
\end{figure*}

\textbf{ Road Places: } Similar to road environment experiments, RGB, RGBS, masked-RGB, S were used to fine-tune a \textit{resnet50} model on places365 for road places. In these frame-based experiments, the approaching, entering and passing sub-classes are treated as separate classes, ie,  \textit{approaching 3-way Intersection}, \textit{entering 3-way Intersection} and \textit{passing 3-way Intersection} are treated as 3 different classes.
In addition to frame-based experiments, standard LSTM\cite{hochreiter1997long} architectures were added to our best frame based models. Such models would allow the capture of the temporal aspect of our labels(\textit{approaching, entering, passing}). While the performance of LSTM and Bi-LSTM models improve our results we hypothesize that decoupling the rough locality \textit{(approaching, entering, passing)} and the event class might help our models to better understand the scene.

Hence we propose a two stream architecture for event proposal and prediction as depicted in Figure \ref{places_block}. The event proposal network proposes candidate frames for the start and end of each event. This involves a classification network to predict \textit{approaching, entering, passing} as the class labels and allows the model to learn temporal cues such as \textit{approaching} is always followed by \textit{entering} and then \textit{passing}. These candidate frames are then sent as an event window to the prediction network. The prediction module aggregates all frames in this window  through global average pooling and produces a singular class label for the entire event. The prediction module is similar to the R2D model in \cite{r2plus1d_cvpr18}. During Testing we first segment out the event windows as proposals, followed by final event classification using  event prediction module.

Summary of our results are shown in Table \ref{roadplacestable}. It must be noted for the temporal experiments, while different input data were used, only our best results of RGB-Masking are displayed. We note that the performance of our model is worse than the BiLSTM model for Branch and Merge classes - possibly because these events are very short and feature averaging done by the prediction module doesn't help.  %\textcolor{red}{ How are instance rsults done. Class wise results are also displayed for our RGB-Frame based experiments, Bi-LSTM and our method in Table \ref{}.As noted blah blah is poor due to blah blah.}  

% The event proposal network uses a resnet50 to extract features which are then sent to an LSTM. A (4 way - Approaching / Entering / Passing / Background) fully connected layer classifies the output of the LSTM to produce the temporal sub-class of the event. 
% The event prediction network is a resnet50 that takes all the candidate frames together and does spatio-temporal Global Average Pooling before sending to the final classification fully connected layer.

%\textcolor{green}{All aforementioned road place classification models can produce only $1$ output at a particular time instance. However, we might have a road place scene where 4-way Intersection occurs with a Zebra-crossing or with a Construction Zone. We note that all road places except Zebra-crossing and Construction Zone are mutually exclusive.}
%To handle these type of situations, we add 2 more final classification fully connected layers whose task is to do binary classification on these classes. We show performance of this Multi-output model on both our best frame-based and temporal models

\begin{table}[]
  \centering
    \begin{tabular}{m{20mm}m{20mm}m{20mm}} \toprule 
        Type            & Input           & Mean F-score \\ \bottomrule
                        & RGB             & 0.208  \\ \cmidrule(l){2-3}
                        & S               & 0.169  \\ \cmidrule(l){2-3}
        Frame Based     & RGBS            & 0.216  \\ \cmidrule(l){2-3}
                        & RGB-Mask        & 0.233 \\  \bottomrule
                        
                        & LSTM             &0.243 \\ \cmidrule(l){2-3}
        Temporal        & Bi-LSTM           &0.275 \\ \cmidrule(l){2-3}
                        & Event Proposal \textbf{(ours}) & 0.285   \\  \bottomrule
    \end{tabular}
   
    \label{roadplacestable}
\end{table}

% \usepackage{multirow}
% \usepackage{graphicx}
\begin{table}[]
\centering
\resizebox{\linewidth}{!}{%
\begin{tabular}{lllllllllllllll}
\hline
\multicolumn{1}{|c|}{\multirow{2}{*}{CLASS}} & \multicolumn{1}{c|}{B} & \multicolumn{3}{c|}{\textbf{I5}} & \multicolumn{3}{c|}{\textbf{RC}} & \multicolumn{3}{c|}{\textbf{C}} & \multicolumn{2}{c|}{\textbf{LM}} & \multicolumn{2}{c|}{\textbf{RM}} \\ \cline{2-15} 
\multicolumn{1}{|c|}{} & \multicolumn{1}{c|}{-} & \multicolumn{1}{l|}{A} & \multicolumn{1}{c|}{E} & \multicolumn{1}{c|}{P} & \multicolumn{1}{c|}{A} & \multicolumn{1}{c|}{E} & \multicolumn{1}{c|}{P} & \multicolumn{1}{c|}{A} & \multicolumn{1}{c|}{E} & \multicolumn{1}{c|}{P} & \multicolumn{1}{c|}{A} & \multicolumn{1}{c|}{P} & \multicolumn{1}{c|}{A} & \multicolumn{1}{c|}{P} \\ \hline
\multicolumn{1}{|l|}{Bi-LSTM} & \multicolumn{1}{l|}{0.88} & \multicolumn{1}{l|}{0.00} & \multicolumn{1}{l|}{0.00} & \multicolumn{1}{l|}{0.09} & \multicolumn{1}{l|}{\textbf{0.24}} & \multicolumn{1}{l|}{0.14} & \multicolumn{1}{l|}{0.46} & \multicolumn{1}{l|}{0.02} & \multicolumn{1}{l|}{0.05} & \multicolumn{1}{l|}{0.29} & \multicolumn{1}{l|}{\textbf{0.09}} & \multicolumn{1}{l|}{\textbf{0.28}} & \multicolumn{1}{l|}{\textbf{0.16}} & \multicolumn{1}{l|}{\textbf{0.23}} \\ \hline
\multicolumn{1}{|l|}{Ours} & \multicolumn{1}{l|}{\textbf{0.92}} & \multicolumn{1}{l|}{0} & \multicolumn{1}{l|}{0} & \multicolumn{1}{l|}{0} & \multicolumn{1}{l|}{0.23} & \multicolumn{1}{l|}{\textbf{0.47}} & \multicolumn{1}{l|}{\textbf{0.46}} & \multicolumn{1}{l|}{\textbf{0.02}} & \multicolumn{1}{l|}{\textbf{0.06}} & \multicolumn{1}{l|}{\textbf{0.38}} & \multicolumn{1}{l|}{0.056} & \multicolumn{1}{l|}{0.08} & \multicolumn{1}{l|}{0.13} & \multicolumn{1}{l|}{0.16} \\ \hline
\multicolumn{1}{|l|}{\multirow{2}{*}{CLASS}} & \multicolumn{1}{l|}{-} & \multicolumn{3}{c|}{\textbf{O/B}} & \multicolumn{3}{c|}{\textbf{I3}} & \multicolumn{3}{c|}{\textbf{I4}} & \multicolumn{2}{c|}{\textbf{LB}} & \multicolumn{2}{l|}{\textbf{RB}} \\ \cline{2-15} 
\multicolumn{1}{|l|}{} & \multicolumn{1}{l|}{-} & \multicolumn{1}{l|}{A} & \multicolumn{1}{l|}{E} & \multicolumn{1}{l|}{P} & \multicolumn{1}{l|}{A} & \multicolumn{1}{l|}{E} & \multicolumn{1}{l|}{P} & \multicolumn{1}{l|}{A} & \multicolumn{1}{l|}{E} & \multicolumn{1}{l|}{P} & \multicolumn{1}{l|}{A} & \multicolumn{1}{l|}{P} & \multicolumn{1}{l|}{A} & \multicolumn{1}{l|}{P} \\ \hline
\multicolumn{1}{|l|}{Bi-LSTM} & \multicolumn{1}{l|}{-} & \multicolumn{1}{l|}{0.23} & \multicolumn{1}{l|}{0.55} & \multicolumn{1}{l|}{0.53} & \multicolumn{1}{l|}{0.03} & \multicolumn{1}{l|}{\textbf{0.28}} & \multicolumn{1}{l|}{\textbf{0.27}} & \multicolumn{1}{l|}{0.14} & \multicolumn{1}{l|}{0.68} & \multicolumn{1}{l|}{0.66} & \multicolumn{1}{l|}{\textbf{0.36}} & \multicolumn{1}{l|}{\textbf{0.22}} & \multicolumn{1}{l|}{\textbf{0.28}} & \multicolumn{1}{l|}{\textbf{0.28}} \\ \hline
\multicolumn{1}{|l|}{Ours} & \multicolumn{1}{l|}{-} & \multicolumn{1}{l|}{\textbf{0.42}} & \multicolumn{1}{l|}{\textbf{0.58}} & \multicolumn{1}{l|}{\textbf{0.59}} & \multicolumn{1}{l|}{\textbf{0.08}} & \multicolumn{1}{l|}{0.16} & \multicolumn{1}{l|}{0.23} & \multicolumn{1}{l|}{\textbf{0.31}} & \multicolumn{1}{l|}{\textbf{0.70}} & \multicolumn{1}{l|}{\textbf{0.67}} & \multicolumn{1}{l|}{0.30} & \multicolumn{1}{l|}{0.19} & \multicolumn{1}{l|}{0.24} & \multicolumn{1}{l|}{0.22} \\ \hline
 &  &  &  &  &  &  &  &  &  &  &  &  &  & 
\end{tabular}%
}
\caption{Summary of F-score results for Places. B-\textit{Background}, I-\textit{Intersection}, RC-\textit{Railway},
C-\textit{Construction}, LM-\textit{Left Merge}, RM-\textit{Right Merge}, LB\textit{-Left Branch}, RB-\textit{Right Branch}, O/B-\textit{Overhead Bridge}}
\label{roadplacestable}
\end{table}

\subsection{Implementation details}
All resent50 models fine-tuned in this paper were pre-trained on the places365 dataset. Data augmentation was performed to reduce over-fitting - random flips, random resize, and random crop were employed. All experiments were performed on NVIDIA P100. All videos were sampled at 3Hz to obtain frames used in experiments. The SGD optimizer was used for frame-based experiments and the Adam optimizer was used for the LSTM based experiments. 

\subsection{Visualization of learned representations}
\begin{figure}
\centering
\subfigure[]{\includegraphics[width=\linewidth]{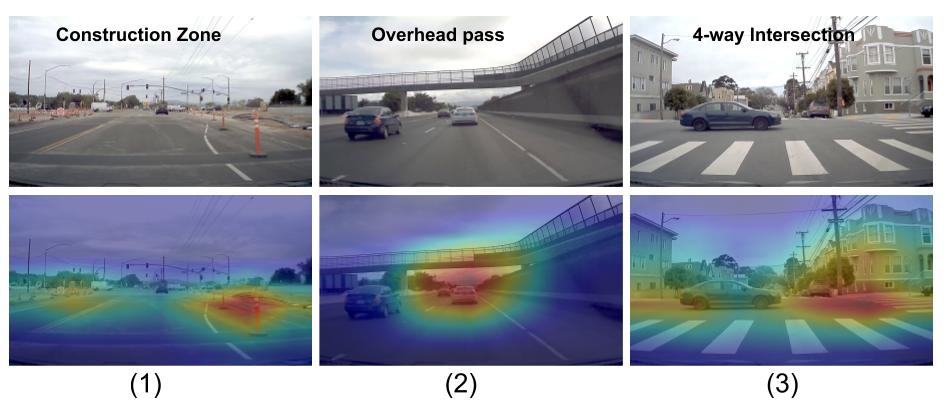}}
\subfigure[]{\includegraphics[width=\linewidth]{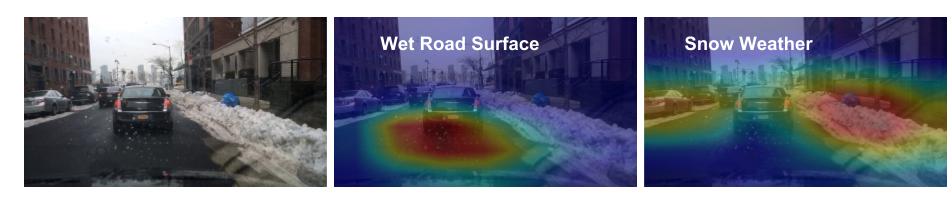}}
\qquad
\caption
{Example localization of our models. (a) Place classification: heat maps show that our models are able to localize the distinctive cues for different place classes,  (b) Weather and Road Surface conditions: the heat-map activations correctly fall on snow regions to predict the weather while always focusing on road to predict the road surface conditions.}
% After masking traffic participants, we observe that the class activation maps (shown on the top right corner) are more robust.
\label{heatmap}
\end{figure}

It has been shown that even CNNs trained on just image labels have localization ability ~\cite{zhou2016learning,selvaraju2017grad,zhang2018top}. Here, we use one such method - Class activation maps ~\cite{zhou2016learning} to show the localization ability of our models. Figure \ref{heatmap} shows some localizations produced by our place, weather and surface classification CNNs.
% \begin{figure}
% %height=6cm,
% \includegraphics[width=\linewidth]{Images/Place-heatmaps.png}
% \centering
% \caption{.}
% \label{placechart}
% \end{figure}

\section{Behavior Understanding}
\begin{figure}
\includegraphics[width=\linewidth]{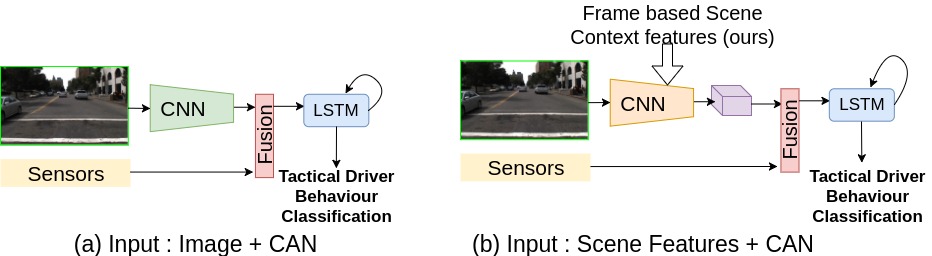}

\centering
\caption{(a) Tactical Driver behavior understanding pipeline in \cite{ramanishka2018toward}. (b) Modification done using our scene context features.}
\label{hddmodel}
\end{figure}

Honda Research Institute Driving Dataset (HDD)~\cite{ramanishka2018toward} was released to enable research on naturalistic driver behavior understanding. The dataset includes 104 hours of driving using a highly instrumented  vehicle.  Ego-vehicle driving behaviors such a \textit{left turn, right turn, lane merge} are annotated. A CNN + LSTM architecture as shown in Figure~\ref{hddmodel} was proposed to classify the behavior  temporally. Due to inclusion of the vehicle's Controller Area Network (CAN) bus signal, which records various signals such as steering angle and speed, the results for \textit{left turn, right turn} were significantly higher than difficult actions such as \textit{lane change, lane merge, and branch}. Hence the current architecture that infers action directly from image fails to capture important cues.

We propose that using an intermediate scene context feature representation helps the model attend to important cues in the scene as evidenced by our attention maps in Figure \ref{heatmap}. For a fair comparison we use the same RGB images to extract intermediate representations from a frame based \textit{resnet50} model trained on our dataset. This would correspond to the first row in our Table  \ref{roadplacestable}. While replacing the input as our scene context features, we keep the rest of the architecture and training protocol exactly the same. Since we only replace the model weights the number of parameters does not change.

As shown in Table~\ref{hddtable}, scene context features improve the overall mean average precision (mAP), especially for rare and difficult classes. Though our model is trained on a different dataset, scene context features embed a better representation as opposed to direct image features. Since the model is able to describe the scene and attend to different regions, it is able to associate actions better with the scenes. For example, lane branch action occurs in the presence of a possible lane branch, or U-Turn generally occurs at intersections; otherwise, it is a false positive. Hence our dataset and pre-trained models can serve as priors (\textit{ false positive removers)} and descriptive scene cues (\textit{soft attention to scene)} for other driving related tasks.

% \begin{figure}
% \centering
% \subfigure[]{\includegraphics[scale=0.5]{Images/HDD1.jpg}}
% \subfigure[]{\includegraphics[scale=0.5]{Images/HDD2.jpg}}
% \qquad
% \caption
% {(a)Tactical Driver behavior understanding pipeline in \cite{ramanishka2018toward}. (b) Modification done using our scene context features.}
% % After masking traffic participants, we observe that the class activation maps (shown on the top right corner) are more robust.
% \label{hddmodel}
% \end{figure}

\begin{table}[t]
\begin{center}
%\small{
\resizebox{0.7\linewidth}{!}{
\begin{tabular}{|*{3}{c|}} 
\hline
Ego-motion behavior & HDD\cite{ramanishka2018toward}& \textbf{Ours}\\ \hline \hline 
Intersection Passing & 0.77 & 0.80\\  \hline 
Left turn & 0.76 & 0.78\\ \hline
Right turn & 0.77 & 0.78\\ \hline
Left lane Change & 0.42 & 0.55\\ \hline
Right lane change & 0.23 & 0.52\\ \hline
Left lane branch & 0.25 & 0.47\\ \hline
Right lane branch & 0.01 & 0.17\\ \hline
Crosswalk Passing & 0.12 & 0.17\\ \hline
RailRoad Passing & 0.03 & 0.02\\ \hline
Merge  & 0.05 & 0.07\\ \hline
U-turn  & 0.18 & 0.29\\ \hline
Overall  & 0.33 & \textbf{0.42}\\ \hline
\end{tabular}
}
\caption{Mean average precision (mAP) without (HDD) and with (Ours) scene context features.}
\label{hddtable}
\end{center}
\end{table}

\section{Conclusion}

In this paper, we introduced a novel traffic scene dataset and proposed algorithms that utilize the spatio-temporal nature of the dataset for various classification tasks.
We demonstrated experimentally that hard attention through semantic segmentation helps scene classification. For the various scene classes studied in this paper, we showed that motion of the scene elements captured by our temporal models provide a more descriptive representations of traffic scene video than simply using their static appearance.

Our models perform better than the conventional CNN + LSTM  architectures used for temporal activity recognition. Furthermore, we have shown the importance of weak object localization for various classification tasks.  Experimental observations based on trained models show that dynamic classification of road scenes provide important priors for higher level understanding of driving behavior.

In future work, we plan to explore annotation of object boundaries for rare classes to achieve better supervision for attention.    Finally, work is ongoing on developing models that are causal and can support multiple outputs to address issues with place classes which are not mutually exclusive (e.g. construction zone at an intersection).

\section*{ACKNOWLEDGMENT}
We would like to thank our colleagues Yi-Ting Chen, Haiming Gang, Kalyani Polagani, and Kenji Nakai for their support and input.

%%%%%%%%%%%%%%%%%%%%%%%%%%%%%%%%%%%%%%%%%%%%%%%%%%%%%%%%%%%%%%%%%%%%%%%%%%%%%%%%

\clearpage
\bibliography{root} 
\bibliographystyle{ieeetr}
%\bibliography{egbib.bib}

\end{document}